\documentclass[runningheads]{llncs}
\usepackage{subcaption}
\usepackage{multicol}
\usepackage{graphicx}
\usepackage{amsmath}
\usepackage{array}
\usepackage{listings}
\begin{document}
\title{Novel Synthetic Data Tool for Data-Driven Cardboard Box Localization\thanks{Supported by the TERAIS project in the framework of the program Horizon-Widera-2021 of the European Union under the Grant agreement number 101079338.}
}
\titlerunning{Cardboard Box Synthetic Data Tool}
%
\author{Peter Kravár\inst{1}\orcidID{0009-0005-8846-9384} \and
Lukáš Gajdošech\inst{2,3}\orcidID{0000-0002-8646-2147}}
\authorrunning{P. Kravár et al.}
%
\institute{
Faculty of Informatics, Masaryk University, Czech Republic \\
\email{xkravar@fi.muni.cz}\\
\and
Faculty of Mathematics, Physics and Informatics, Comenius University, Slovakia\\ 
\email{gajdosech@fmph.uniba.sk}
\and
Skeletex Research, Slovakia
}
\maketitle              
\begin{abstract}
Application of neural networks in industrial settings, such as automated factories with bin-picking solutions requires costly production of large labeled data-sets. This paper presents an automatic data generation tool with a procedural model of a cardboard box. We briefly demonstrate the capabilities of the system, its various parameters and empirically prove the usefulness of the generated synthetic data by training a simple neural network. We make sample synthetic data generated by the tool publicly available.
\end{abstract}

\section{Introduction}

Automatic detection and localization of bins on a conveyor belt is an essential task in automated factories. This detection must be robust to guarantee the safe operation of robotic arms. It includes handling edge cases such as missing edges, occlusion, and variance in the materials and shapes of the bins. Moreover, in a specific scenario of package delivery factories, bins are made from a non-rigid cardboard material. These boxes are prone to various deformations, and their paper flaps are semi-randomly opened while being filled by workers and robots.  

Analytical detection algorithms lack robustness and are hard to modify for new cases \cite{Katsoulas2003}. On the other hand, machine learning based methods require data. Capturing real RGB-D samples in various scenarios in factories is costly. Therefore, the generation of synthetic data is recently a popular research topic \cite{Periyasamy2021,Chen2022}, outlined by the boom of commercial solutions such as NVIDIA Omniverse™ \footnote{https://www.nvidia.com/en-us/omniverse/solutions/digital-twins/}.

Following our previous work \cite{Gajdosech2021}, in this short submission we propose a novel data-generation tool for the automated generation of training data containing cardboard boxes. We evaluate the results of a neural network trained upon this novel data against a baseline synthetic generator, which has no automatic parametrization and cannot produce boxes with paper flaps. 

\section{Generating Data}

This project aimed to create a high-level system for generating synthetic datasets of 3D bin scans using Blender 3D compiled into a python module \texttt{(bpy)}\footnote{https://docs.blender.org/api/current/info\_advanced\_blender\_as\_bpy.html}. We accomplished this by wrapping Blender's functionality into high-level classes representing respective parts of the 3D scanning pipeline. Our pipeline simulates the real scanning process of a structured light scanner. Render settings, scanner parameters and the behavior of random parameter generation are fully customizable by the user. The output of our system comes in the form of structured point cloud data.
The camera transformation matrix and the volume box of the generated cardboard box are also exported and used as ground truth data. 




\subsection{Parametric Cardboard Box}
\begin{figure}
    \centering
    \begin{subfigure}[b]{0.49\textwidth}
        \centering
        \includegraphics[width=0.9\columnwidth]{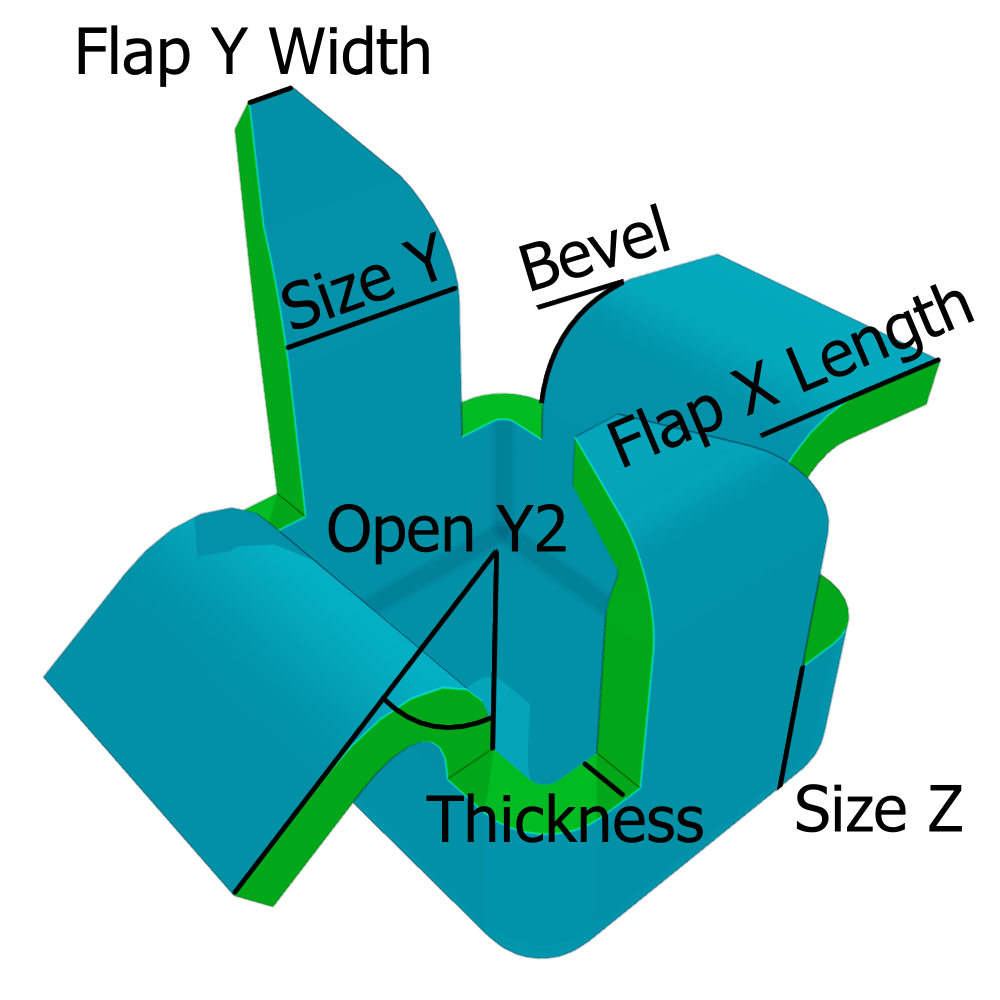}
        \caption{parameters of the model}
        \label{fig:params}
    \end{subfigure}
    \begin{subfigure}[b]{0.49\textwidth}
        \centering
        \includegraphics[width=0.9\columnwidth]{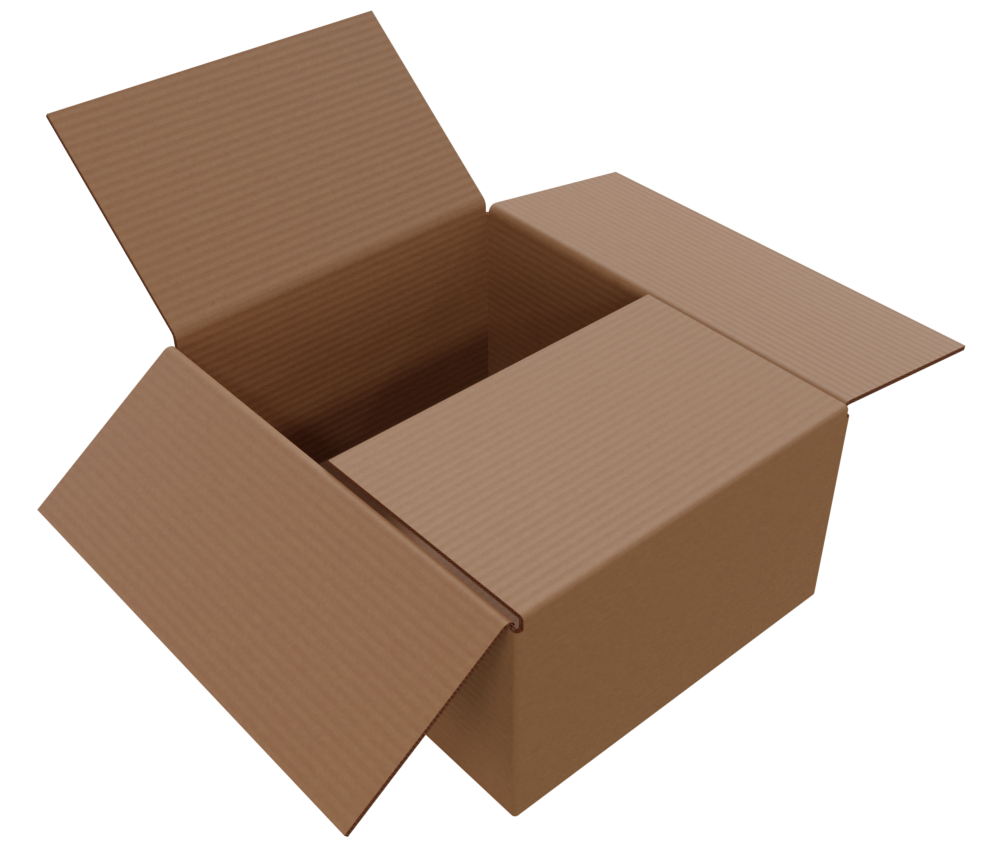}
        \caption{procedurally shaded cardboard box}
        \label{fig:shaded_box}
    \end{subfigure}
    \caption{Illustration of box parameters and the resulting rendered image.}
    \label{fig:boxes}
\end{figure}

Variety in synthetic data can be achieved by randomizing parameters of appropriate parametric model \cite{Fedorova2021}. For the purpose of generating virtual cardboard boxes, we have created a parametric model which approximates the most significant box features, see Figure \ref{fig:params} for visual illustration. By changing the parameters, we are able to obtain a wide variety of virtual cardboard boxes. The box parameters are:
\begin{multicols}{2}
    \begin{enumerate}
        \item \textbf{Size} - box dimensions
        \item \textbf{Flap Length} - flap dimensions
        \item \textbf{Flap Width} - flap taper
        \item \textbf{Open} - flap open angle
        \item \textbf{Thickness} - cardboard thickness
        \item \textbf{Bevel} - roundness of box edges
    \end{enumerate}
\end{multicols}

\vspace{-12mm}

\begin{figure}
    \centering
    \includegraphics[width=2.3cm]{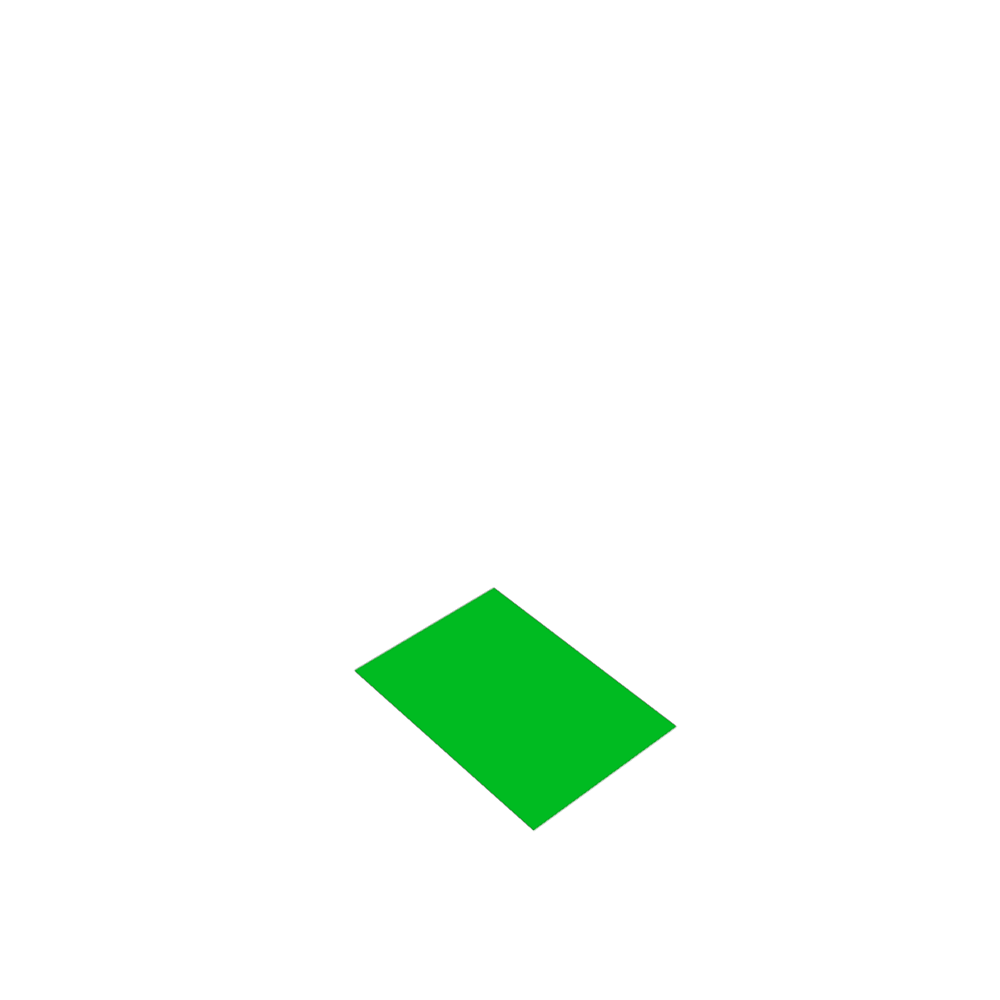}
    \includegraphics[width=2.3cm]{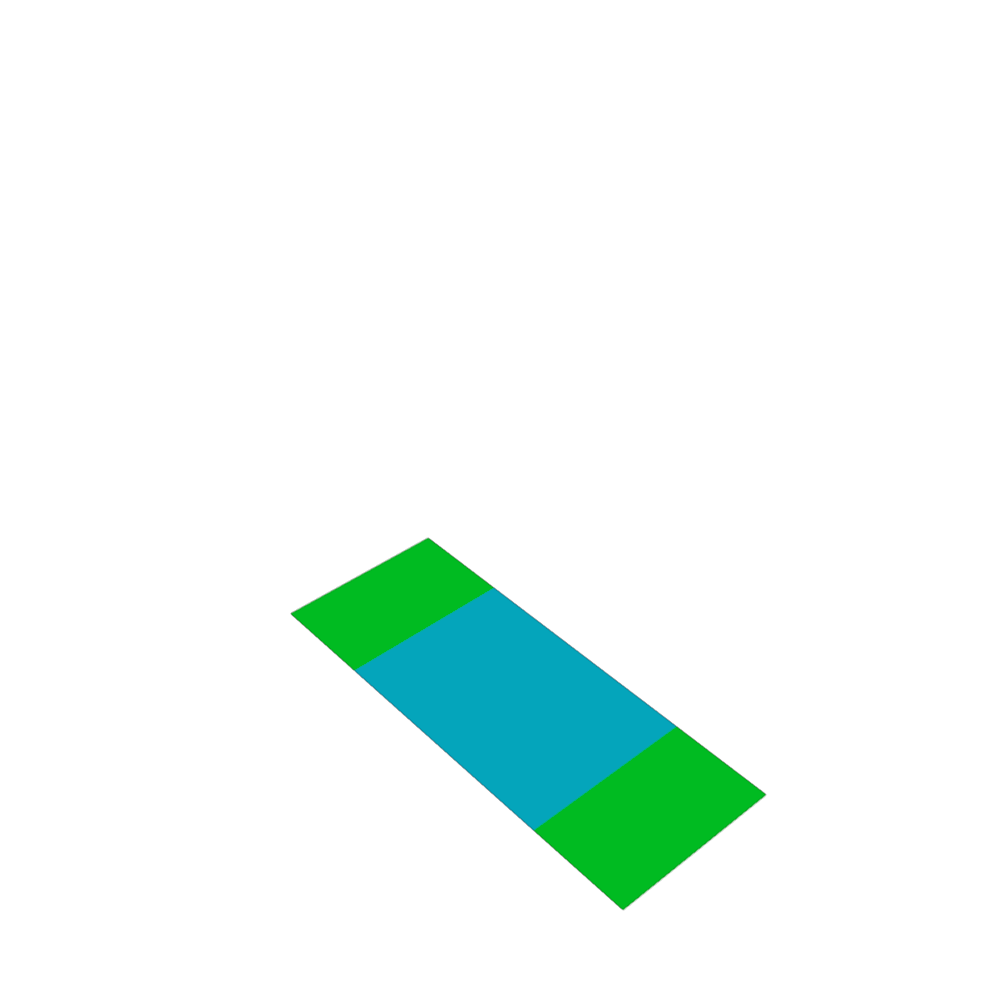}
    \includegraphics[width=2.3cm]{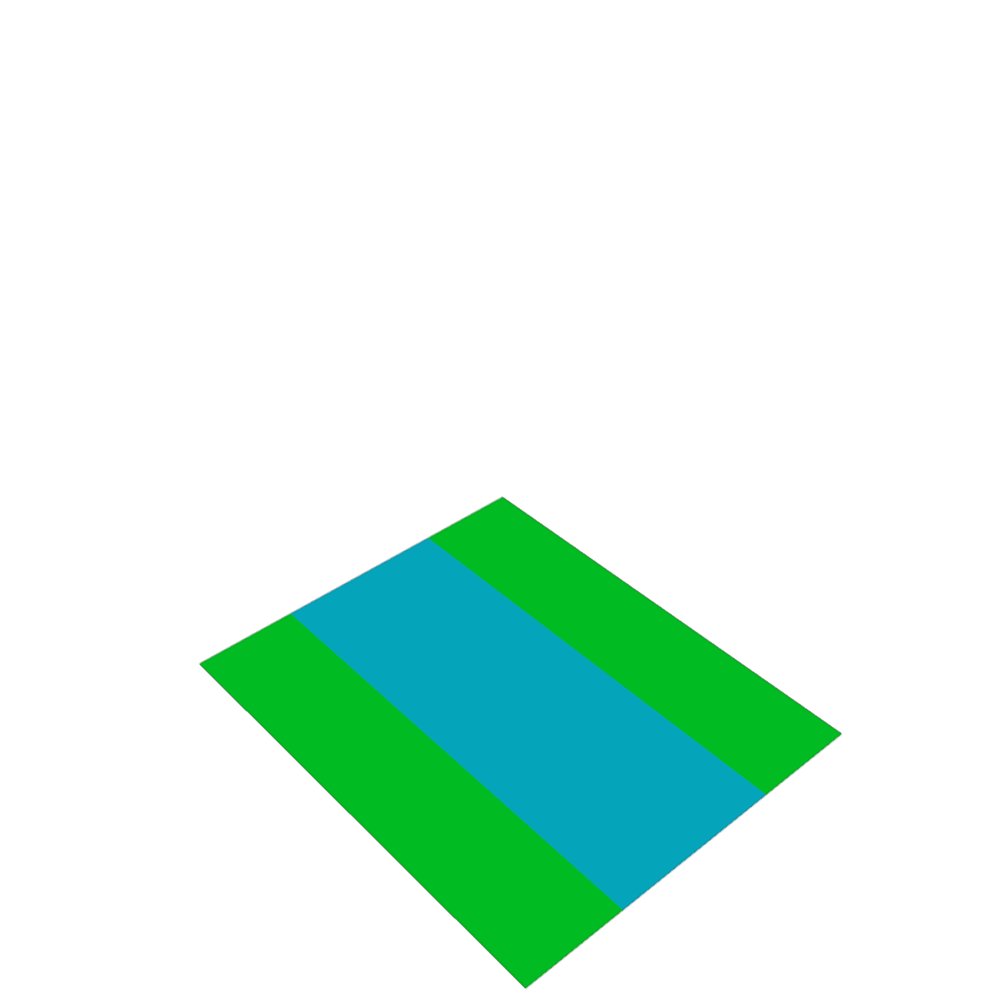}
    \includegraphics[width=2.3cm]{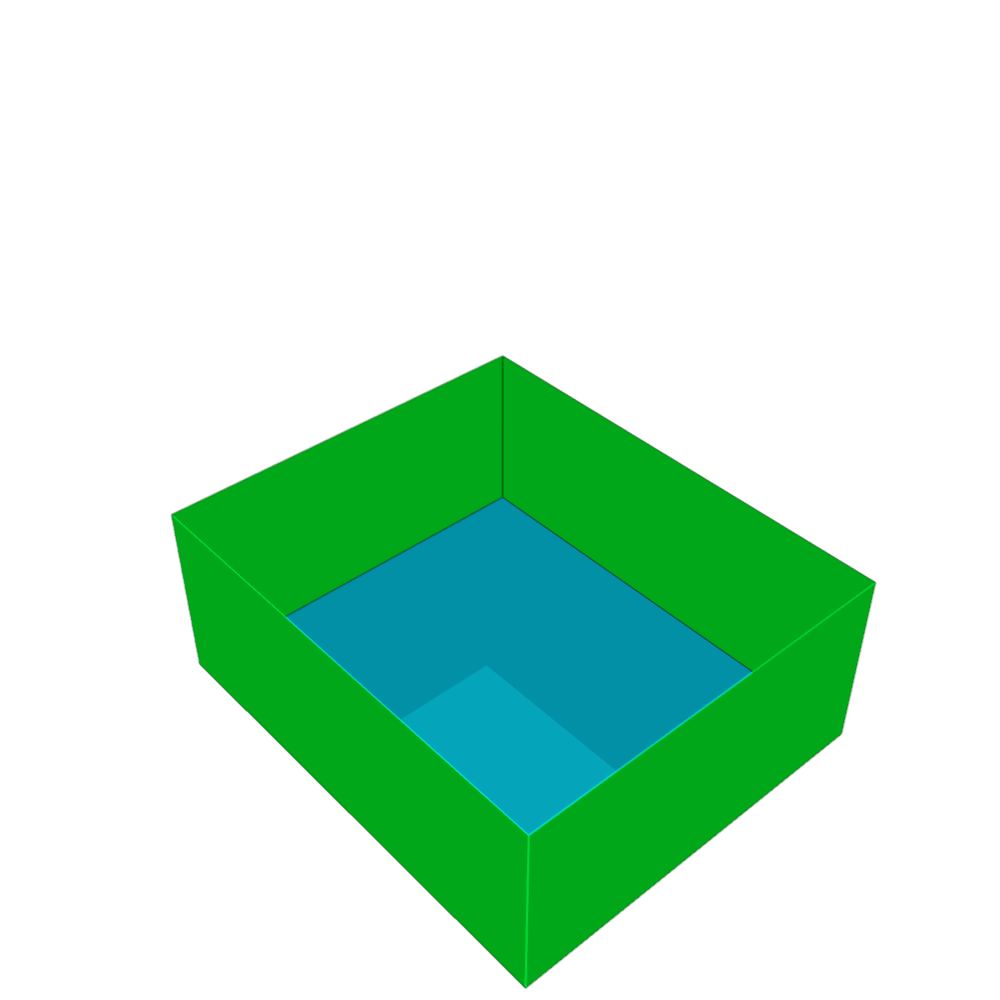}
    \includegraphics[width=2.3cm]{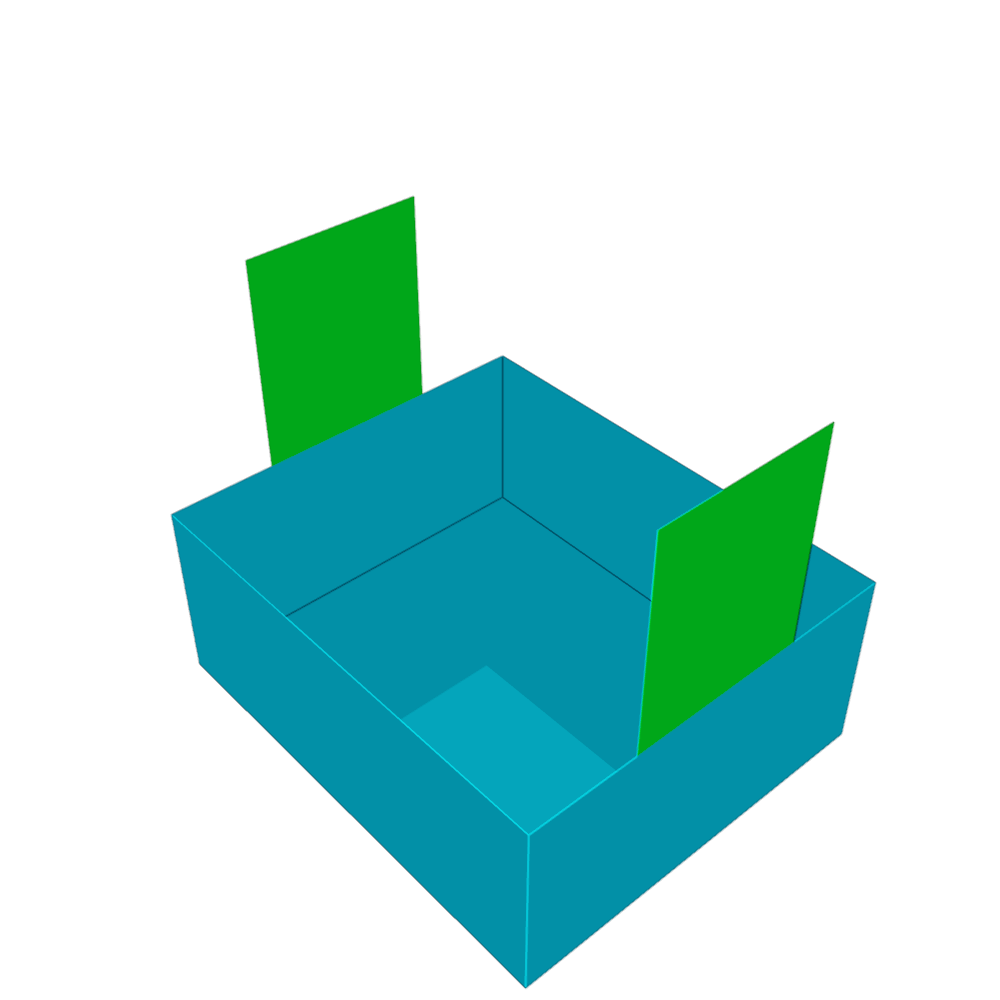}
    \includegraphics[width=2.3cm]{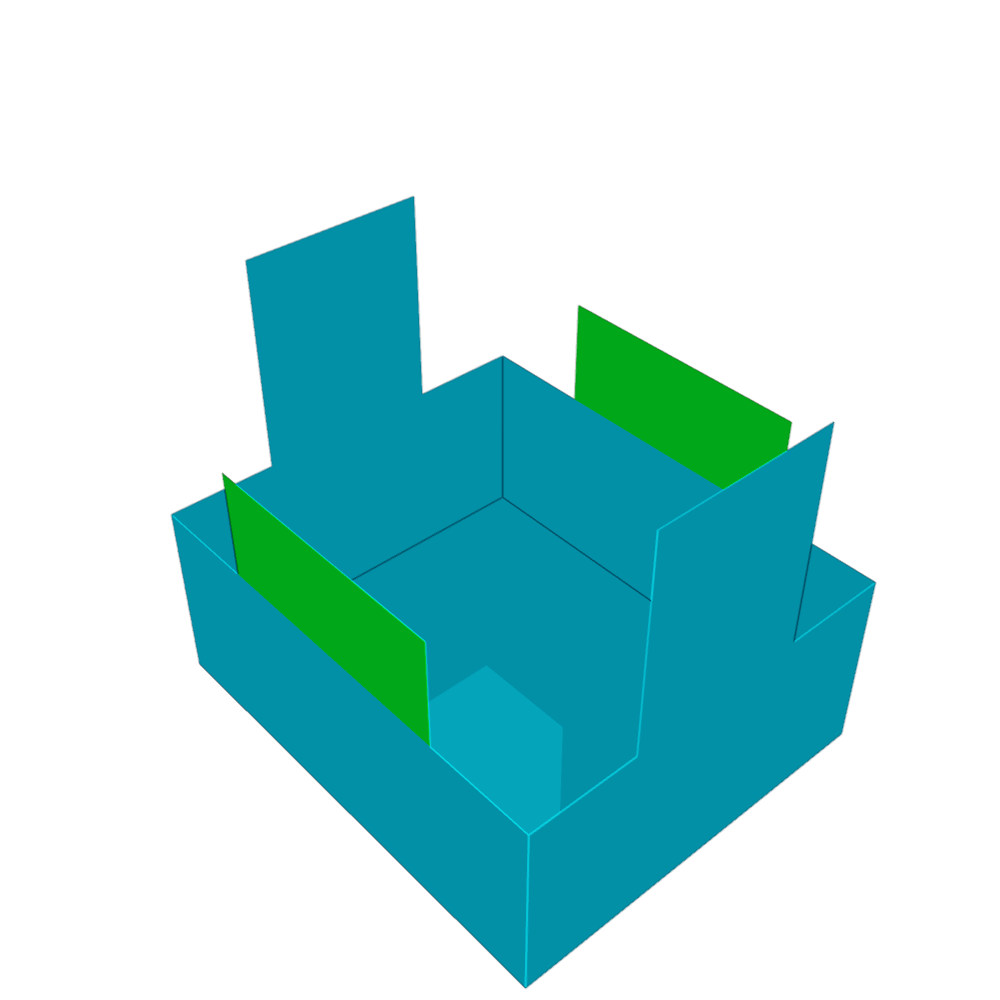}
    \includegraphics[width=2.3cm]{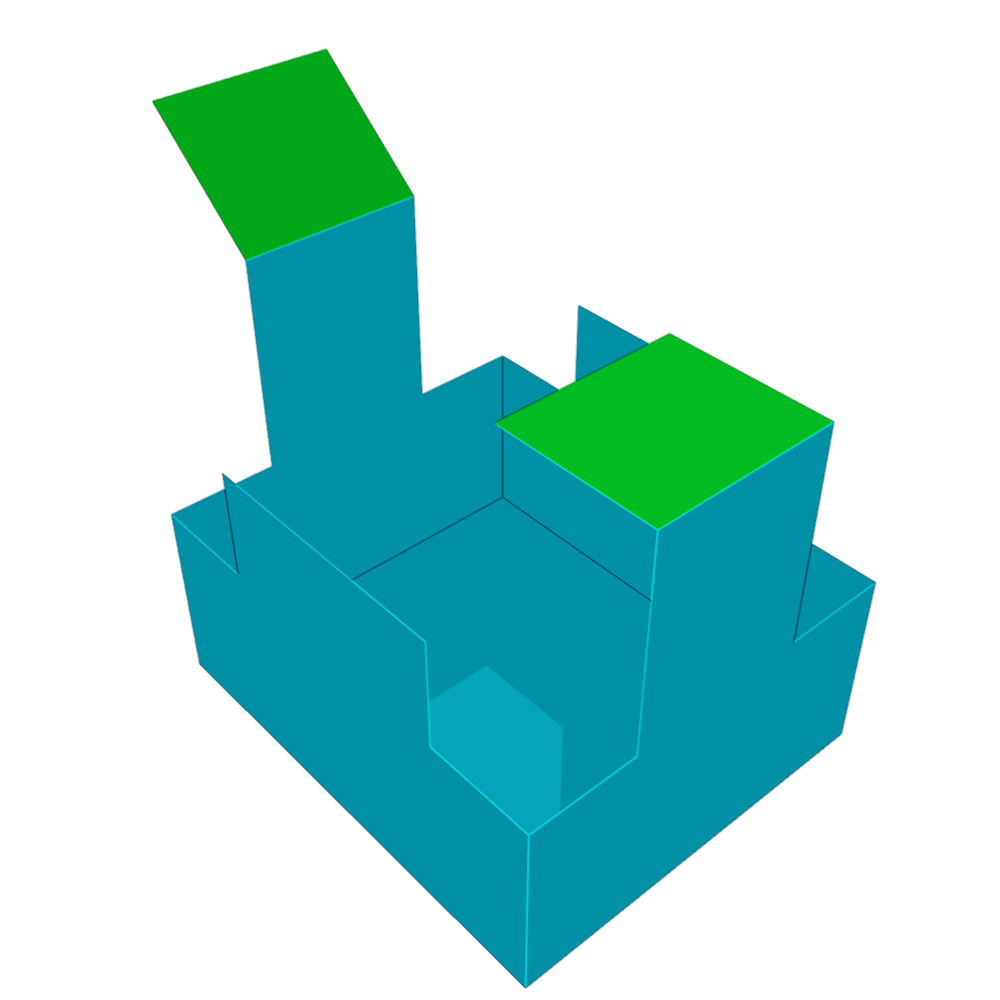}
    \includegraphics[width=2.3cm]{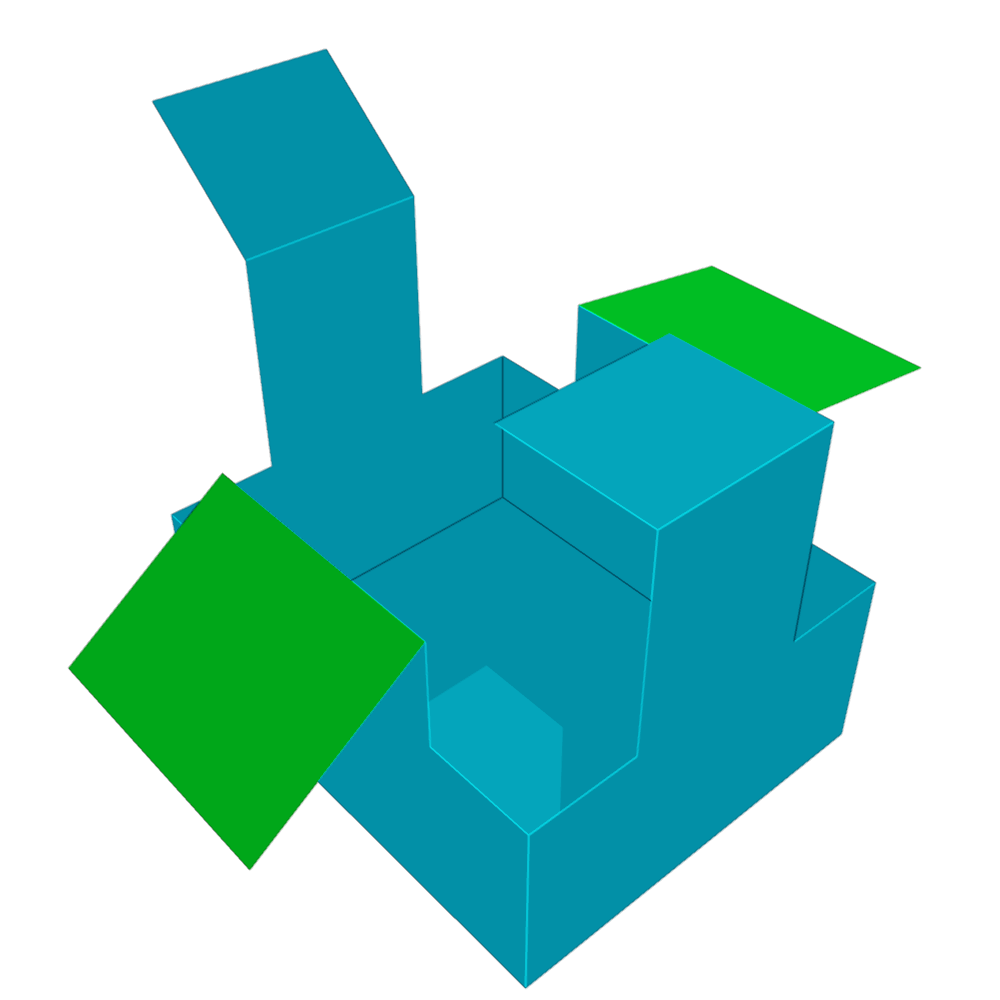}
    \includegraphics[width=2.3cm]{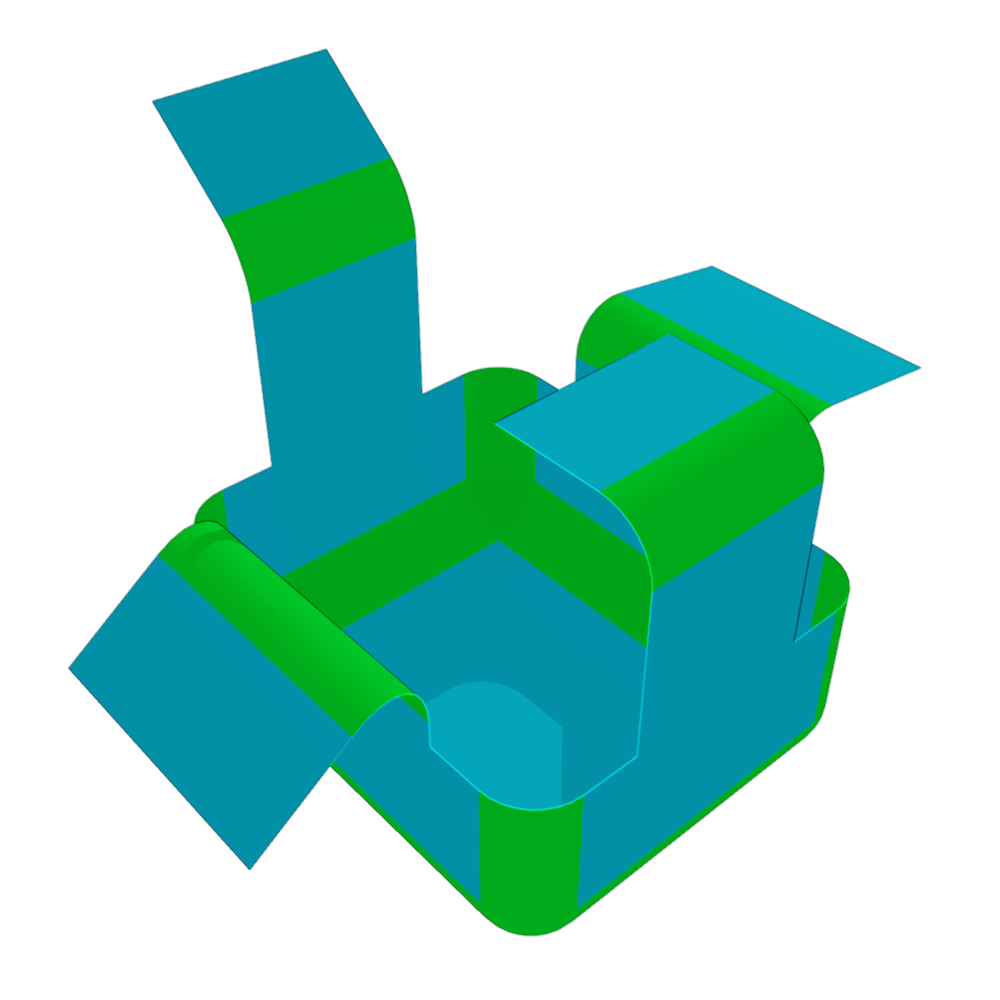}
    \includegraphics[width=2.3cm]{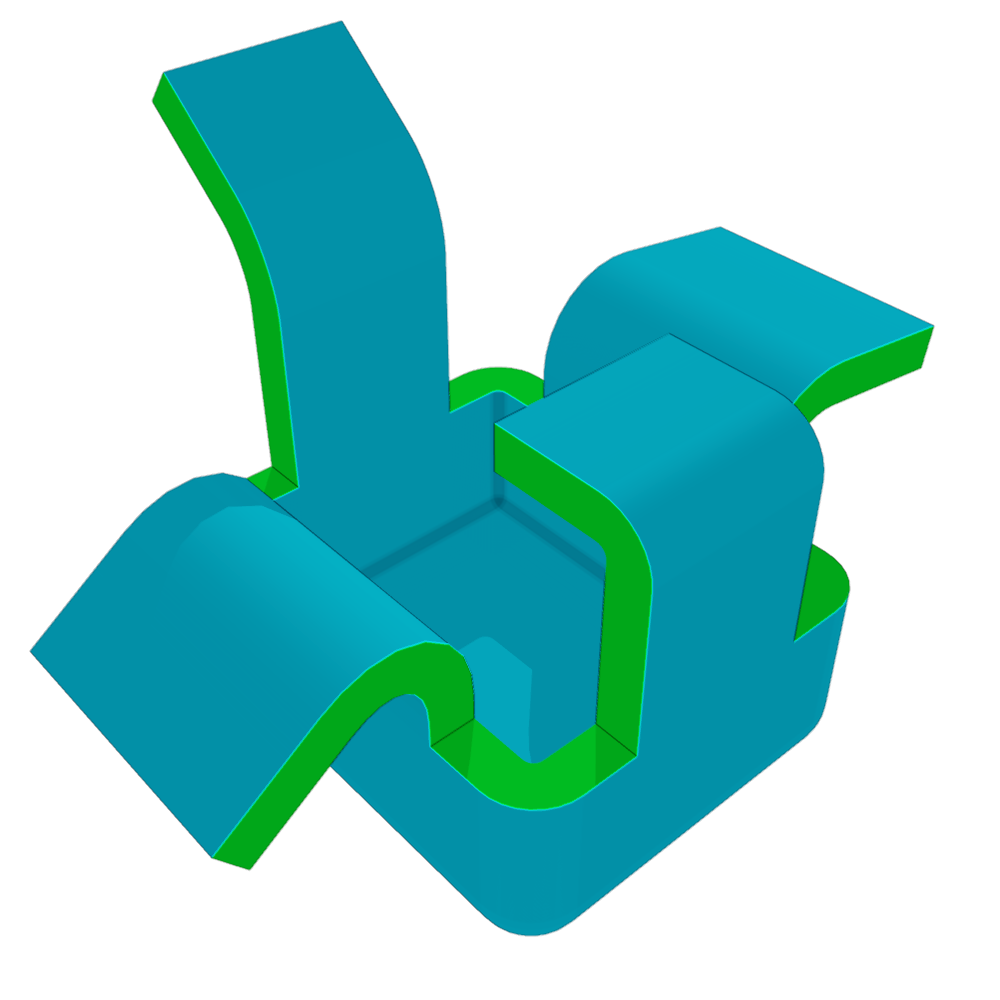}
    \caption{Box creation process, operations are exaggerated for visual clarity.}
    \label{fig:box_creation}
\end{figure}

\vspace{-5mm}

We have approximated a generic cardboard box as an object created using corresponding sequence of steps as shown in Figure \ref{fig:box_creation}. The steps include a series of extrusions followed by rounding corner edges and adding thickness. The parametric model is implemented using Blender's Geometry Nodes system \cite{Gumster2022}.

In real production, a box is assembled by folding a sheet of cardboard. The resulting object can therefore be closely approximated in 2D. Such 2D representation can serve as a UV map without visible seams, used for procedural shading of the parametric cardboard box, 
Figure \ref{fig:shaded_box} shows the resulting rendered image.

\subsection{Generation Parameters}
\label{parameter_randomness}
The camera location was generated as a random unit vector in the positive \texttt{XYZ} part of a sphere scaled by uniformly distributed random distance in the $(1m, 1.7m)$ interval. The rotation of the scanner was then calculated such that the camera would point at world origin. Generation of boxes utilized random distributions for multiple parameters, ex. a single dimension was randomized as: 

$$
\text{Size}_X = 0.25 + min(max (- \sigma \times \gamma, \; \mathcal{N}(\mu,\,\sigma^{2})), \sigma \times \gamma).
$$

\noindent
For our experiments, we set $\sigma = 0.1$ and $\gamma = 2.0$, each constant is in SI units.





\section{Experiment}

We have verified the added value of the proposed generator by training a neural network for 6D pose estimation of the cardboard boxes \cite{Gajdosech2021}. We have created two sets of synthetic training data, each consisting of 496 samples. The first set was generated using a baseline generator, without the automated box parametrization and flaps, see Figure \ref{fig:bingen}. The second set is generated using our novel tool. The data, together with loading scripts in python is available at our website\footnote{https://www.skeletex.xyz/portfolio/datasets}.

 \subsection{Metrics}

Translation of the box origin is evaluated using euclidean distance: $e_\mathrm{TE}(\, \hat{\vec{t}}, \vec{t} \,) = \lVert\,\vec{t} - \hat{\vec{t}}\,\rVert_2$. For rotation, we use model-independent angle distance between rotational axes calculated from corresponding rotation matrices as:
$
e_\mathrm{RE}(\hat{R}, R) = \min_{\hat{R'} \in \{\hat{R_1}, \hat{R_2}\}} \mathrm{arcos} ((\mathrm{Tr}(\hat{R'}R^{-1})-1)/2),
$
where $\mathrm{Tr}$ is the matrix trace operator.

\subsection{Evaluation}

\begin{figure}
    \centering
    \begin{subfigure}[b]{0.32\textwidth}
        \centering
        \includegraphics[width=1.0\columnwidth]{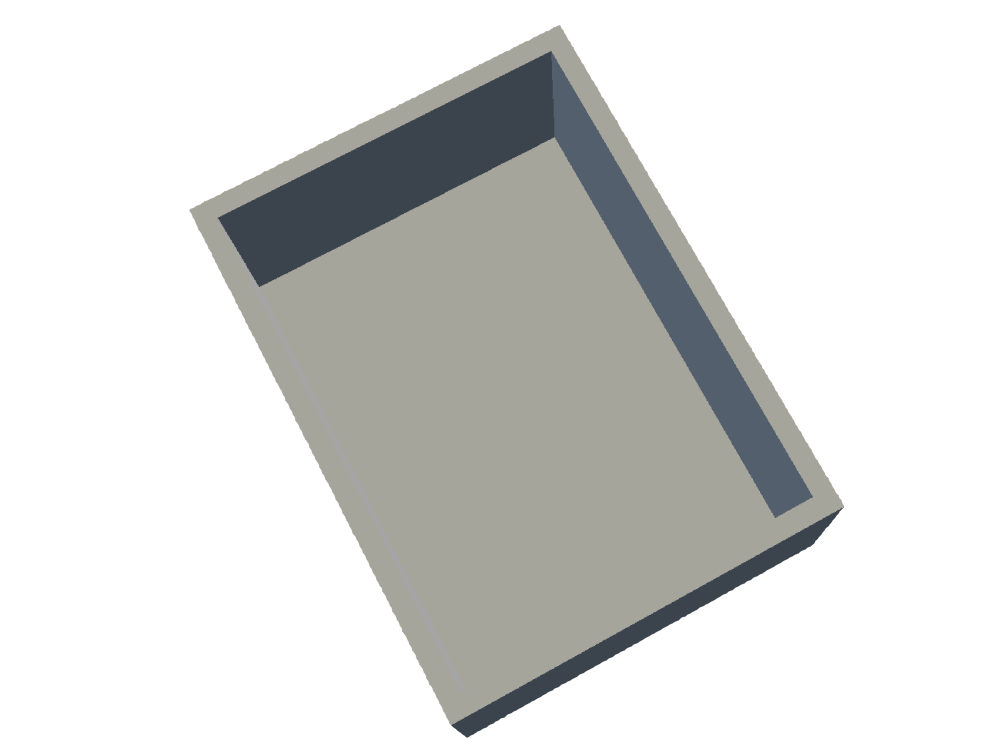}
        \caption{baseline generator}
        \label{fig:bingen}
    \end{subfigure}
    \begin{subfigure}[b]{0.32\textwidth}
        \centering
        \includegraphics[width=1.0\columnwidth]{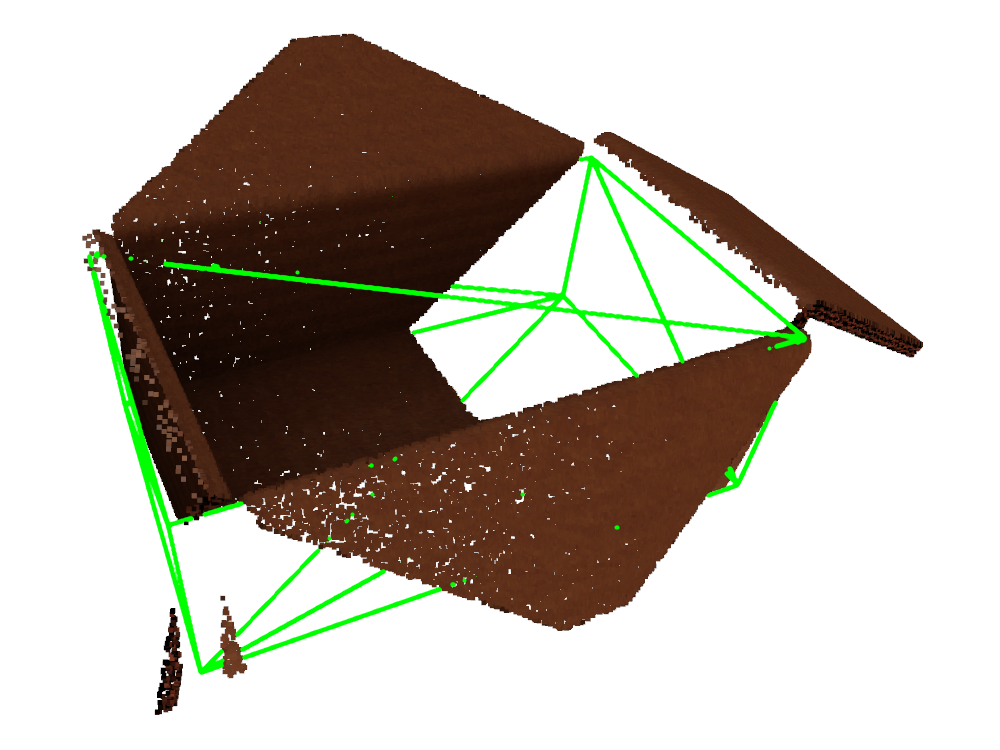}
        \caption{validation sample}
        \label{fig:val}
    \end{subfigure}
    \begin{subfigure}[b]{0.32\textwidth}
        \centering
        \includegraphics[width=1.0\columnwidth]{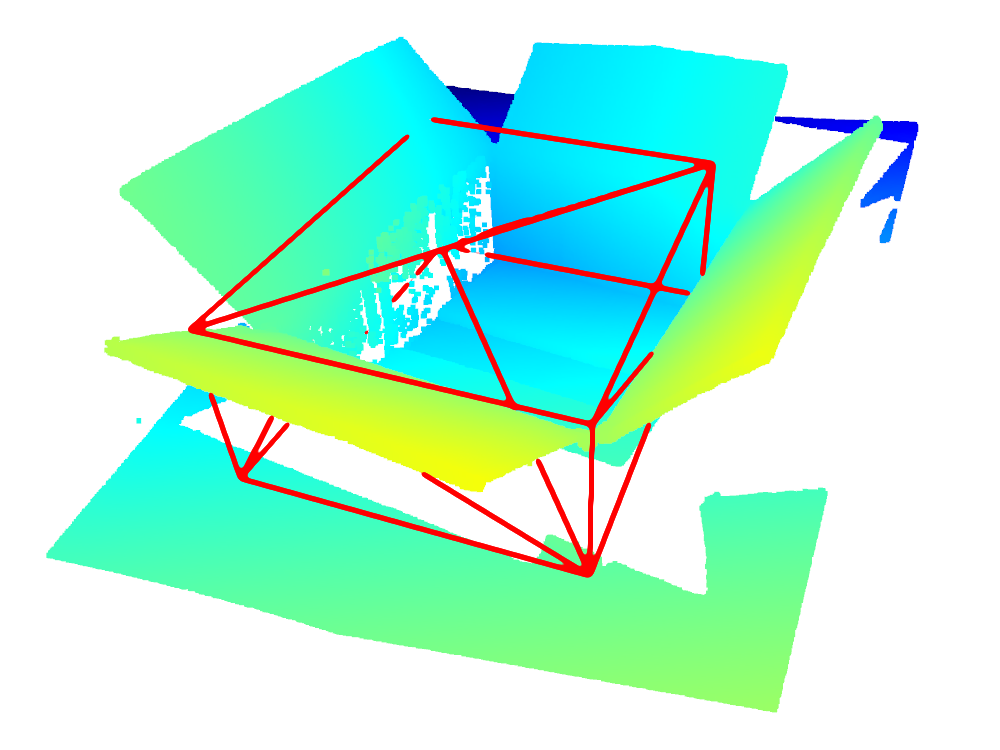}
        \caption{test sample}
        \label{fig:test}
    \end{subfigure}
    \caption{Sample from the baseline generator and network predictions.}
    \label{fig:compare}
\end{figure}

Table \ref{tab:results} compares networks trained over the two synthetic datasets. The validation set consists of $100$ synthetic samples from the proposed generator and test set of $22$ real samples captured by PhoXi 3D Scanner\footnote{https://www.photoneo.com/phoxi-3d-scanner/}. Figure \ref{fig:compare} shows qualitative examples of the network predictions. Note that it has only the 3D point cloud on the input, without any information about the dimensions of the boxes. 

We conclude that the novel generator helped the network to generalize and learn to ignore paper flaps, showing promise in improving synthetic-data tools for more successful training. Future work includes expanding this tool for additional possible variances, such as bins from semi-transparent plastic materials with a simulation of physical phenomena like light caustics in photo-realistic textures.  

{\renewcommand{\arraystretch}{1.2}
\begin{table}[t]
\centering
\caption{Comparison of network's performance using different training data.}
\label{tab:results}
\begin{tabular}{p{30mm}||c|c|c|c}
\hline
Training Data & val $\overline{e_\mathrm{TE}}$ (mm) & val $\overline{e_\mathrm{RE}}$ (rad) & test $\overline{e_\mathrm{TE}}$ (mm) & test $\overline{e_\mathrm{RE}}$ (rad) 
\\
\hline
\hline
Baseline Synthetic & 35.603 & 1.336 & 14.237 & 1.121 \\
\hline
Novel Synthetic & 4.326 & 0.240 & 12.161 & 0.787 \\
\hline
\end{tabular}
\end{table}
}

\vspace{-11mm}

%
%
%
\bibliographystyle{splncs04}
\bibliography{mybibliography}
%





\end{document}